\documentclass[default]{sn-jnl}

\usepackage{amsmath,amssymb,amsfonts}
\usepackage{graphicx}
\usepackage{subfigure}
\usepackage{float}
\usepackage{soul}
\usepackage{array}
\jyear{2021}%

\newcolumntype{M}[1]{>{\centering\arraybackslash}m{#1}}

\theoremstyle{thmstyleone}%
\theoremstyle{thmstyletwo}%

\theoremstyle{thmstylethree}%

\begin{document}

\title[An OPG based FPAD using GAN]{An Open Patch Generator based Fingerprint Presentation Attack Detection using Generative Adversarial Network}


\author*[1]{\fnm{Anuj} \sur{Rai}}\email{phd1901201003@iiti.ac.in}

\author[1]{\fnm{Ashutosh} \sur{Anshul}}\email{ashutoshanshul01@gmail.com }

\author[1]{\fnm{Ashwini} \sur{Jha}}\email{ashwini.jha186@gmail.com }
\author[1]{\fnm{Prayag} \sur{Jain}}\email{offpjain@gmail.com }
\author[2]{\fnm{Ramprakash} \sur{Sharma}}\email{ram.psharma28@gmail.com }
\author[1]{\fnm{Somnath} \sur{Dey}}\email{somnathd@iiti.ac.in }

\affil[1]{\orgdiv{Computer Science and Engineering}, \orgname{Indian Institute of Technology Indore}, \orgaddress{\street{Simrol}, \city{Indore}, \postcode{453552}, \state{Madhya Pradesh}, \country{India}}}

\affil[2]{\orgdiv{Computer Science and Engineering}, \orgname{The LNM Institute of Information Technology}, \orgaddress{\street{Sumel}, \city{Jaipur}, \postcode{302031}, \state{Rajasthan}, \country{India}}}


\abstract{
The low-cost, user-friendly, and convenient nature of Automatic Fingerprint Recognition Systems (AFRS) makes them suitable for a wide range of applications. This spreading use of AFRS also makes them vulnerable to various security threats. Presentation Attack (PA) or spoofing is one of the threats which is caused by presenting a spoof of a genuine fingerprint to the sensor of AFRS. Fingerprint Presentation Attack Detection (FPAD) is a countermeasure intended to protect AFRS against fake or spoof fingerprints created using various fabrication materials. In this paper, we have proposed a Convolutional Neural Network (CNN) based technique that uses a Generative Adversarial Network (GAN) to augment the dataset with spoof samples generated from the proposed Open Patch Generator (OPG). This OPG is capable of generating realistic fingerprint samples which have no resemblance to the existing spoof fingerprint samples generated with other materials. The augmented dataset is fed to the DenseNet classifier which helps in increasing the performance of the Presentation Attack Detection (PAD) module for the various real-world attacks possible with unknown spoof materials. Experimental evaluations of the proposed approach are carried out on the Liveness Detection (LivDet) 2015, 2017, and 2019 competition databases. An overall accuracy of 96.20\%, 94.97\%, and 92.90\% has been achieved on the LivDet 2015, 2017, and 2019 databases, respectively under the LivDet protocol scenarios. The performance of the proposed PAD model is also validated in the cross-material and cross-sensor attack paradigm which further exhibits its capability to be used under real-world attack scenarios.}




\maketitle

\section{Introduction}\label{sec1}
Fingerprint biometric systems are beneficial compared with conventional approaches as they are trustworthy, autonomous, and user-friendly. These properties make them useful in many real-life applications such as border security, national identification, and various payment systems. Nowadays, we can also spot these systems in various banking applications for customer identification and authentication such as Automated Teller Machines (ATMs), customer account opening through biometric-based Aadhar identification, etc.  Its applicability in such a wide range of applications makes it vulnerable to various security threats, such as theft of identity, account hacking, unauthorized access, and many more. A Presentation Attack (PA) is one of the possible threats, which is carried out by presenting an artificial artifact of a genuine user's fingerprint to the sensor to get access to the system. Nowadays, various spoofing materials including rubber fingers, 3D printed papers, and high-definition images and videos of fingerprints that are capable to deceive a  system are available at a very low cost. Gelatine, Wood Glue, Latex, and Siligum are some of the inexpensive materials available to create an artificial replica of a genuine user's fingerprint. This attack necessitates the use of altered fingerprints, as well as fingerprint patterns that have been edited or deleted, to influence the identification process. This problem of PAs can be addressed with two kinds of methods i.e. hardware-based and software-based methods. The hardware-based methods utilize extra hardware devices for the extraction of features such as temperature, humidity, and pulse rate from the fingertip. The utilization of additional hardware costs in terms of money, as well as maintenance, makes these approaches less popular for Presentation Attack Detection (PAD). On the other side, software-based methods rely only on the input fingerprint samples for PAD which makes these approaches user-friendly as well as cost-effective. Due to the widespread applicability and cost-effectiveness of the software-based approaches, our work is focused on developing a software-based approach to deal with the PAs in the fingerprint recognition system.
\par

The state-of-the-art software-based approaches include statistical feature-based methods, perspiration and pore based-methods, handcrafted feature-based methods, and deep learning-based methods. The perspiration property of the fingerprint is affected by the pressure that is applied to the fingertip as well as the temperature of the surrounding environment. Extraction of these feature values requires multiple impressions of the same fingerprint. Pore-based methods require a sensing device with the capability of capturing high-resolution (around 1000 pixels per inch) images of fingerprints which proportionally increases the cost of the biometric system. The existing feature-based methods are comprised of a variety of statistical and handcrafted features. The application of different sensing devices causes significant variations in the quality of fingerprint samples which affects the performance of these methods. This phenomenon can be seen in the fingerprint images captured using the orcanthus sensor. The poor quality of images in the orcanthus dataset makes it difficult to extract accurate hand-crafted features from them. These methods are capable to detect the spoof fingerprint in the intra-sensor and intra-material experimental scenarios but do not resemble the same performance in the cross-sensor or cross-material paradigm. Various researchers have adopted deep learning-based systems to deal with this problem due to their superior image classification capability. These models have a series of convolutional layers that extracts minute features from fingerprint images. Although, deep learning is a strong tool, as proved in \cite{b4_uliyan}\cite{b18} but most of these methods are not capable of dealing with the problem of FPAD in cross-material and cross-sensor paradigms due to less generalizable capability across different sensor and materials. The main objective of the proposed method is to empower a deep-learning classifier in such a way that makes it capable to detect the spoof samples fabricated with unknown material as well as captured with some unknown sensors.

\par
In this paper, we propose a deep learning-based approach that outperforms various existing methods for FPAD. Since detection of fingerprint samples fabricated with unknown spoof materials i.e. open-set PAD is a key issue for the security of fingerprint recognition systems, we have developed a unique generator called an Open-Patch Generator (OPG) that enables a classifier to deal with the problem of open-set as well as cross-sensor PAs. We have used the Generative Adversarial Network (GAN), which works on the unsupervised learning method of machine learning. GAN consists of two sub-models that are generator and discriminator. The generator creates fingerprint samples, while the discriminator determines if the input sample is real or created from the generator model. One more benefit of using GAN to create these samples is that the generated sample does not belong to a particular class of samples while properties of input samples are present in it. Since the utilization of GAN models costs in terms of time and computational resources, we have used it for the generation of spoof samples only. A game-play between both these sub-models enables the generator to create more realistic fingerprint samples which are further utilized to train DenseNet classifiers. DenseNet is a deep CNN architecture in which each layer is feed-forward connected to the other layers. The feature maps from all previous layers are fed as input to each layer, and the layer's feature maps are fed as input to the succeeding layers. We have used DenseNet as a classifier due to its capability to deal with the problem of vanishing gradient. This problem persists with the input images that have less feature or texture information. 
\section{Related Work}
PAs are one of the major security threats to Automatic Fingerprint Recognition Systems (AFRS). To protect the AFRS against PAs, various methods are proposed in the literature. In recent years, various software-based methods are proposed by various researchers to solve the problem of PAs in AFRS. According to the methodology and utilization of resources, these methods can be further divided into categories such as statistical feature-based methods, perspiration and pore-based methods, handcrafted feature-based methods, and deep learning-based methods. Some of the methods belonging to these categories as well as their limitations are mentioned in this section. \\
\textbf{Statistical feature-based methods:}
Choi et al. \cite{b7}, suggested a novel fingerprint liveness detection method based on statistical features such as histogram, directional contrast, ridge thickness, and ridge signal. These features are combined and used for the training of the SVM classifier on a customized fingerprint database.
Park et al. \cite{b11}, utilized statistical features such as deviation, variance, skewness, kurtosis, hyper skewness, and hyper flatness along with three additional features i.e. average brightness, standard deviation, and differential image, to train an SVM classifier for the liveness detection on ATVS Fake Fingerprint database. Their method have not been tested in cross-material and cross-sensor paradigms.
Marasco et al. \cite{b25} proposed a feature-based solution by focusing on some static characteristics of fingerprint images. The texture of an image provides some visual information related to the variation of grey-level intensity and orientation. They retrieved certain first-order statistical features from the fingerprint image, such as energy, entropy, median, variance, etc., as well as some intensity-based features to train different classifiers on LivDet 2009 database. Statistical feature-based methods are quite sufficient to be utilized as an important feature but none of these methods have shown the capability to detect spoofs in cross-material and cross-sensor paradigms. Therefore, the problem remains unsolved for the novel spoof materials and sensors by these methods.
\\
\textbf{Perspiration and pore-based methods:}
Pores are small holes that are present in live fingerprints. This property is utilized by researchers to measure the liveness of a fingerprint sample.
Derakshini et al. \cite{b27} utilize the prevalence of perspiration pattern and sweat diffusion pattern using ridge signal extracted from fingerprint samples to discriminate between live and spoof fingerprints.
Espinoza et al. \cite{b16} demonstrate the use of pores for the detection of PAs. Sweat pore is a natural pattern that persists in human fingers and remains missing in spoof fingerprints. Authors have utilized one of the pore properties such as number of pores as a feature for the detection of spoof fingerprints. The proposed method is validated using a custom-made database. Abhyankar et al. \cite{b19} proposed a wavelet-based method for the detection of fingerprint PAs. They exploited the sweat feature of fingerprints to discriminate between live and spoof samples. Sweat is a natural property that is not found in spoof fingerprints and hence it plays a vital role in discrimination between live and spoof fingerprint samples. Marcialis et al. \cite{b17} proposed a pore analysis method for the detection of PAs. The system captures two fingerprint impressions at an interval of five-second and then detects the number of pores in them. The proposed work is evaluated using a custom-made database consisting of 8960 live and 7760 spoof fingerprint images.\\
The major disadvantage with pore features as a liveness detector is that it can only be detected in an image that is captured with a high-resolution sensor while the presence of wetness or perspiration features in a fingerprint depends on the atmosphere and external temperature. Also, these methods are not capable to detect a spoof fingerprint sample in cross-material and cross-sensor paradigms.\\
\textbf{Handcrafted feature-based methods:}
Dubey et al. \cite{b1} proposed a shape and texture feature-based method. In their study, Speeded Up Robust Feature [SURF] and Pyramid Extension of Histogram of Gradient [PHOG] are utilized to extract shape information which works as a distinctive feature for the detection of PAs. The ridges and valleys present in live and spoof fingerprints have different shapes due to the different elasticity levels possessed by the human skin and fabrication materials. Along with the above-mentioned features, the gabor wavelet is used to analyze the texture information. They evaluated the proposed method on LivDet 2011 and 2013 databases. The proposed method performs well in intra-sensor but have not been tested in cross-material and cross-sensor paradigms.
Rattani et al. \cite{b2} proposed a novel method to detect PAs against the samples generated with the help of unknown materials. They suggested a novel material detector which is a multi-class classifier to predict a fingerprint as a live, spoof, or unknown. The proposed method have shown better performance for the limited set of fingerprints but it have not been tested on any of the available standard fingerprint databases.
In another work, Rattani et al. \cite{b3} suggested the use of a Weibull-calibrated SVM as a classifier. This SVM is a combination of 1-class SVM and binary SVM. This modification has shown a significant improvement in performance as compared to their previous work.
Sharma et al. \cite{b5} used quality-related features such as ridge-valley smoothness, number of abnormal ridges and valleys, ridge-valley clarity, frequency-domain analysis, orientation-certainty level, and gabor quality which play a vital role in the detection of spoof fingerprints. The live and spoof fingerprints have a different range of values for these features due to the natural properties of live fingerprints which are not present in their spoofs. The proposed method is validated on the LivDet 2009, 2013, and 2015 databases. In continuation to \cite{b5}, Sharma et al. \cite{anshulA} proposed to extract white circular patches, ridge ending, ridge bifurcation, and center intensity difference for spoof detection. The proposed model is evaluated on LivDet 2015 databases. Galbally et al. \cite{b6} suggested 25 novel features for liveness detection in three authentication systems i.e. face, iris, and fingerprint. The proposed work have shown good performance in the intra-sensor paradigm.
Ghiani et al. \cite{b12} utilized a local image descriptor called Binary Statistical Image Feature (BSIF). This feature is obtained by a set of natural filters on fingerprint images whose output is then converted to binary data. The proposed method is tested on Livdet 2011 database but not tested in the cross-material and cross-sensor paradigm. In another work, Sharma et al. \cite{b32} utilized various hand-crafted texture descriptors such as Weber Local Descriptor (WLD), Binary Statistical Image Features (BSIF), Local Phase Quantization (LPQ), Local Contrast Phase Descriptor (LCPD), and Rotation Invariant Co-occurrence among Adjacent Local Binary Pattern (RICLBP) for FPAD. The results on LivDet 2011, 2013, and 2015 databases shows that none of the texture descriptors performs uniformly well for the cross-sensor, cross-material experimental scenarios. Xia et al. \cite{b13} suggested a gradient-based method that extracts the second and third order co-occurrence matrix of gradients and utilized it as a feature for the training of the SVM classifier. This method is tested on LivDet 2009 and 2011 databases.
In another work, Xia et al. \cite{b20} proposed a local descriptor that extracts intensity variance as well as gradient-based features. The co-occurrence probability of these two features is assessed to form a feature vector which is further used to train an SVM classifier. This method performs well as compared to their previous approach when validated using LivDet 2011 and 2013 databases but not tested in open-set and cross-sensor paradigms.
Gragnaniello et al. \cite{b22} also utilizes the weber descriptor to extract digital excitation and gradient information. Despite calculating the co-occurrence matrix, it computes a joint histogram of them to build feature vectors. This method is validated using LivDet 2011 and 2013 databases.
Yuan et al. \cite{b33} proposed a method that relies on the creation of two co-occurrence matrices using the laplacian operator. This operator is used to compute image gradient values for different quantization operators. These matrices are utilized as a feature for the training of the back-propagation neural network. This method is validated using LivDet 2013 database.
Kim et al. \cite{b34} proposed a novel image descriptor that utilizes the local coherence of a given fingerprint image. The creation of spoof fingerprints using different fabrication materials results in a significant difference between fingerprints. This method uses a local coherence pattern as a feature to train SVM. The proposed method is validated using ATVS and LivDet 2009, 2011, 2013, and 2015 databases. Sharma et al. \cite{b37} have suggested a hybrid approach to deal with the problem of PAs. They suggested a novel feature named Local Adaptive Binary Pattern (LABP) which is an enhanced version of the existing Local Binary Pattern (LBP). They have combined this feature with existing BSIF and Complete Local Binary Pattern (CLBP) and used them for the training of the SVM classifier to generate a global liveness score for fingerprint samples. The proposed method is validated on LivDet 2009, 2011, 2013, and 2015 databases. 
The methods discussed in this category perform well for the intra-sensor and material scenarios but do not maintain the same performance in cross-sensor and cross-material scenarios. \\
\textbf{Deep learning-based methods:}
Arora et al. \cite{b26}, proposed a robust framework to detect PAs in fingerprint biometric systems which involves contrast enhancement using histogram equalization as a pre-processing technique. They have used VGG architecture as a classifier. Authors have validated this work on a variety of fingerprint databases which include FVC 2006, ATVSFFp, finger vein database, LivDet 2013, and 2015 databases.  
Uliyan et al. \cite{b4_uliyan} proposed a deep features-based spoof detection method. It utilizes a Deep Boltzmann Machine [DBM] for the extraction of features from fingerprint images that is capable to find a complex relationship among the features. The proposed method exhibits better performance when compared with hand-crafted feature-based methods but suffers from poor performance in cross-sensor and cross-material PAD scenarios.
Gomez-Barrero et al. \cite{b38} proposed a multi-modal architecture which relies on the surface of the fingerprint within the shortwave infrared region of the Short Wave Infra Red (SWIR) sensor as well as Laser Speckle Contrast Image (LSCI) technology. The SWIR sensor captures an RGB image of the fingerprint which is classified by Resnet and VGG19 classifiers. In addition to this, averaged LSCI image undergoes through some feature extraction modules and then the extracted features are used by SVM for liveness score generation. Scores generated by both models are fused to get a global liveness score.
Chugh et al. \cite{b15} proposed a deep learning-based method that uses minutiae-centered fingerprint images for the training and testing of a MobileNet CNN classifier. A fingerprint is cropped into a finite number of patches based on the number of minutiae points. The extracted patches are fed to a MobileNet CNN model which generates a liveness score for every patch. The score produced by the classifier for all the patches are fused together to compute global score. This method have been tested on LivDet 2011, 2013, and 2015 databases as well as the Michigan State University's FPAD database.
In continuation to their previous work, Chugh et al. \cite{b14} proposed another method for the detection of spoofs created using unknown fabrication materials. Since, novel and cheap materials are being discovered nowadays, it is hard to generalize an FPAD model for all kind of spoofs. They have utilized an image synthesis technique mentioned in \cite{b14b} to create new fingerprint samples which contribute to better training of deep learning classifier. The method is  tested on LivDet 2015 and 2017 databases.
Nogueira et al. \cite{b18}, utilized  pre-trained CNN architectures using style transfer. Their method involves various architectures such as VGG, Alexnet, and CNN with SVM. The method is tested over LivDet 2009, 2011, and 2013 databases.

The above study concludes that deep learning-based methods outperform the methods belonging to other categories. The possession of convolutional layers enables them to extract discriminating features. Some of the proposed methods \cite{b26}, \cite{b4_uliyan}, and \cite{b18} performs well for the LivDet databases but still their performance is not well for the PAD in cross-material and cross-sensor paradigms.

\section{Proposed Work}
In this paper, an OPG wrapper that is composed of a GAN is proposed for the detection of PAs in fingerprint recognition systems. The proposed wrapper is capable to generate fingerprint samples of other possible unknown spoof materials.
The proposed method comprises three steps: 1. Training of Wasserstein's GAN model as an OPG wrapper, 2. Generating spoof fingerprint patches using the OPG wrapper and, 3. Training the section-wise DenseNet classifiers with training dataset along with the spoof fingerprint patches created by the OPG wrapper. The liveness score of a fingerprint image is obtained by the fusion of scores generated for all section-wise patches.
In the following sections, descriptions of the GAN, Wasserstein's GAN, OPG, and DenseNet classifier are provided.

\subsection{\textbf{ Generative Adversarial Network (GAN)}}
The GANs are a combination of two models, the generator and the discriminator. Both the generator and the discriminator models are trained to complement each other such that their performances will be improved in the subsequent epochs. The generator takes the noise vector as input and generates new data samples, whereas the discriminator predicts the possible image source S (real images from the training dataset or OPG-generated samples) of the given input sample. The log-likelihood function for predicting the image source S is given in Eq. \ref{eqn:gan_log_like}. Here, $E[X]$ denotes the expectation value of variable $X$, $P$ is the probability distribution over image source S, $X_{real}$ denotes an image from input dataset and $X_{fake}$ denotes an image generated by the generator.
\begin{equation}
\label{eqn:gan_log_like}
L = E[log P(S = real \| X_{real})]+ E[log P(S = fake \| X_{fake})]
\end{equation}
The discriminator model of the GAN tries to maximize the above equation while the generator model tries to minimize the second term of the above equation.

\subsection{\textbf{ Wasserstein’s GAN with Gradient Penalty}}
Although GANs are proven to be successful generators, it always remained a challenging problem to train them. Therefore, to make the generators more efficient, various techniques \cite{WGAN4}\cite{WGAN2}\cite{WGAN1}\cite{WGAN3}\cite{WGAN5} suggested different generator objectives for training GANs.  Arjowsky et al. \cite{WGAN5} have examined the convergence properties of the GAN optimized value functions. Arjowsky et al.  \cite{WGAN1}, further, proposed a new variation of basic GAN called Wasserstein's GAN which uses Earth-Mover's distance as a loss function in place of Binary Cross Entropy Loss. Basic GANs suffer from the problem of vanishing gradient and mode collapse which are removed by using Earth-Mover's distance as a loss function. In this architecture, the generator tries to maximize the distance between the probability distributions for real samples as well as generated samples while the discriminator tends to minimize this distance. However, W-GANs sometimes may fail to converge or generate low-quality images due to weight clipping. Gulrajani et al. \cite{WGAN_GP} introduced gradient penalty as an alternative to weight clipping which overcomes the shortcomings present in W-GAN.

\subsection{\textbf{Open Patch Generator (OPG)}}
The purpose of devising an OPG is to generate spoof fingerprint patches to improve CNN's learning capabilities against the samples of unknown spoof materials and unknown sensors. This process involves the training of separate W-GAN models for every sensor. A W-GAN for a particular sensor A is trained with spoof fingerprint patches captured from every sensor except sensor A so that no data from the testing sensor gets involved in the training process. For instance, training a W-GAN corresponding to HiScan includes spoof fingerprint patches from all the sensors available in the training dataset used, except HiScan. The wrapper is finally used to generate new generalized spoof samples. This method helps the model in generalized training as the models are trained with samples from multiple sensors and materials. The protocol of using the OPG is associated with the testing sensor. For example, devising a model to be tested on HiScan will require the use of patches generated from W-GAN corresponding to HiScan from the OPG. Since the training sample used to train W-GAN corresponding to HiScan doesn’t include samples from the same sensor, this protocol ensures that no sample from the testing domain is a part of the training, thus, maintaining zero correlation between training and testing datasets. A more detailed explanation is presented while discussing intra-sensor and cross-sensor experimental scenarios in section \ref{compAnalysis}.

\subsection{\textbf{ DenseNet}}
\begin{figure*}[]
	\centering
	\resizebox{1.13\textwidth}{!}{
		{
			
			\includegraphics[]{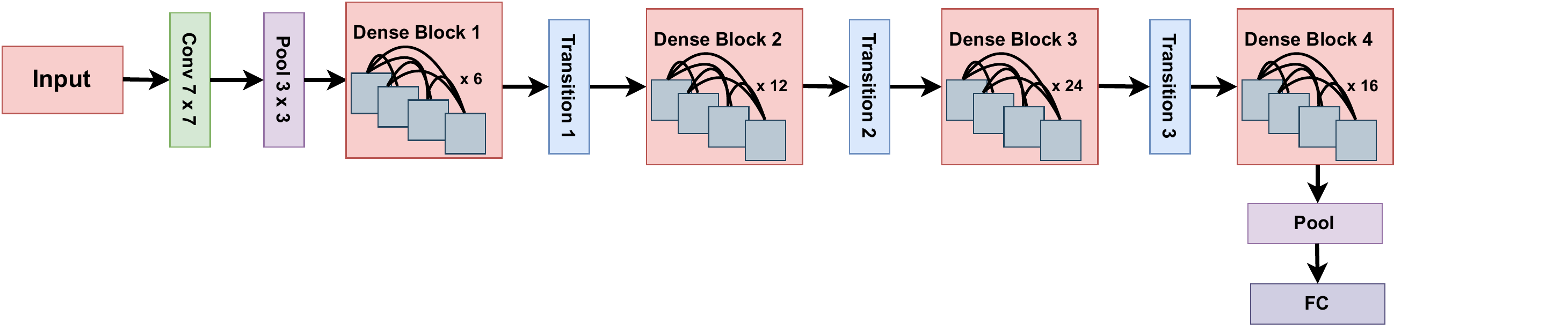}
	}}
	\caption{Architecture of DenseNet-121 classifier}
\vspace{3mm}	
\label{DenseNet}
\end{figure*}

\begin{figure*}[t!]
	\centering
	\resizebox{0.27\textheight}{!}{
		{
			
			\includegraphics[]{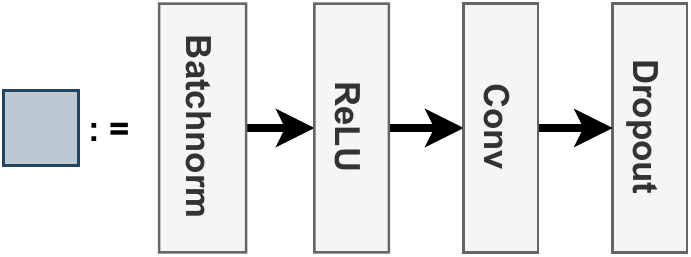}
	}}
	\caption{Composition of dense block}
\vspace{3mm}	
\label{DenseNet2}
\end{figure*}

\begin{figure*}[t!]
	\centering
	\resizebox{0.33\textheight}{!}{
		{
			
			\includegraphics[]{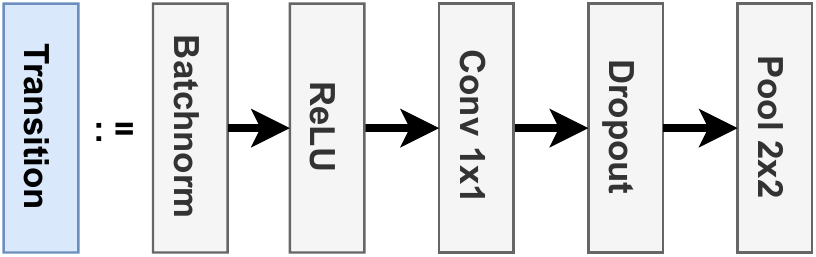}
	}}
	\caption{Composition of transition block}
\vspace{3mm}	
\label{DenseNet3}
\end{figure*}

DenseNet \cite{b9} is a CNN model in which each layer is connected directly with every other layer. The traditional CNN architectures suffer from the problem of vanishing gradient. A vanishing gradient is an issue that appears in deep CNNs when their depth is increased. Due to the involvement of a large number of activation functions, the gradient of the loss function approaches zero. As fingerprint images have limited texture and color information as compared with imagenet datasets, CNNs face the problem of vanishing gradient while being applied for classification on these images. The dense connections among the layers as well as batch-normalization operation in the transition block of the DenseNet classifier reduce the problem of vanishing gradient and make it suitable to be used for fingerprint PAD.\\
The DenseNet model (version-121) is comprised of four dense blocks. Each block has 6, 12, 24, and 16 dense layers as shown in Fig. \ref{DenseNet}. The internal architecture of densenet is depicted in Fig. \ref{DenseNet2}. Each dense layer consists of three sequential operations including Batch-Normalization, Rectified Linear Unit (ReLU), and Convolution. Thus it is defined by BN - ReLU - Conv where Conv indicates convolutional layers with a $3 \times3 $ kernel. Each layer collects feature maps from all preceding layers. The output feature map can be denoted with Eq. \ref{eqn:densenet}.
\begin{equation}
\label{eqn:densenet}
X_n = F_n[A_0, A_1, A_2, A_3, .....A_{n_-1}] 
\end{equation}
where $A_0, A_1, A_{n-1}$ represents the concatenation of all feature maps from layers 0, 1, 2, ..., n-1. $F_{n}$, denotes a function that performs BN -\> ReLU -\> Conv ($3 \times 3$). In addition, a bottleneck layer is used after each dense block to perform 1x1 convolution with $2 \times 2$ average pooling. As a result, the densenet 121 model completely replicates the procedure as: BN -$>$ ReLU -$>$ Conv($3 \times $3) -$>$ BN-$>$ ReLU -$>$ Conv($1 \times 1$)...n-times. The transition block consists of one more operation which is pooling. The internal architecture of the transition block is depicted in Fig. \ref{DenseNet3}.

\vspace{4 mm}

\vspace{-3mm}
\subsection{\textbf{Training}}
In the training phase, first, the proposed OPG is trained to create new fingerprint patches belonging to the spoof class. The generated spoof patches along with the training dataset are used for the training of the DenseNet CNN classifier. The entire process is described in the following sections.
\subsubsection{\textbf{ Minutiae-centered patch extraction}}
The fingerprint minutiae points (ridge bifurcation and ridge endings) are detected using the National Institute of Standards and Technology's Biometric Image Software (NBIS) \cite{NBIS}. NBIS requires input images to be in .jpg format, the training and testing images are first converted to the .jpg format and then fed to the NBIS. We have extracted minutiae-centered patches instead of random patches as minutiae are the most important features for fingerprint matching. One more advantage of using patches in this way is that it makes the data uniform irrespective of the different dimensions of fingerprint images captured with different types of sensors. The output of the mindtct package of NBIS includes detected minutiae points, each represented by four values-
(i) x-coordinate ($x_{i}$) value, (ii) y-coordinate ($y_{i}$) value, (iii) orientation angle ($\theta_{i}$), and (iv) quality ($q_{i})$ ranging from 0 to 100. The quality value output helps to filter out low-quality minutiae points by setting up a quality threshold.

\begin{figure}[t]
	\centering
	\subfigure[threshold - 0 \hspace{5 mm} (c) threshold - 15]
	{	
		\includegraphics[width=0.15\textwidth]{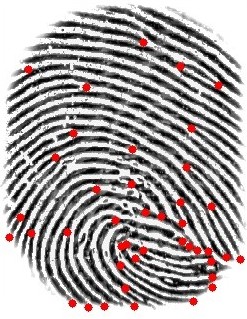}
		\hspace{8 mm}
		\includegraphics[width=0.15\textwidth ]{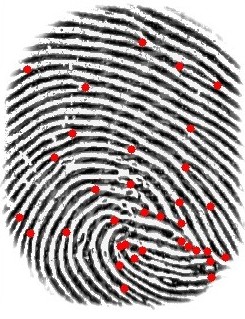}
	}
	\vspace{4 mm}
	
	\subfigure[threshold - 10 \hspace{4 mm} (d) threshold - 20]
	{
		\includegraphics[width=0.15\textwidth]{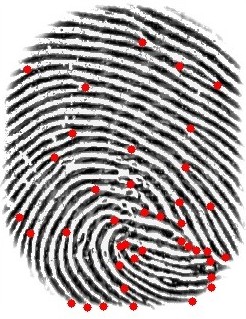}
		\hspace{8 mm}
		\includegraphics[width=0.15\textwidth]{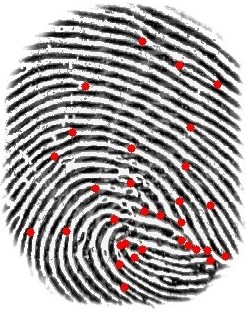}
	}
	\vspace{4mm}
	\caption{Detected minutiae points for quality threshold values (a) 0, (b) 10, (c) 15, (d) 20}
	\label{QualityThresh}
	\vspace{-2mm}
\end{figure}

We selected minutiae points for a set of sample fingerprint images using a quality threshold varying between 0 to 20. Fig. \ref{QualityThresh} shows the selected minutiae points at quality thresholds of 0, 10, 15, and 20. In Fig. \ref{QualityThresh}, it can be seen that the threshold value of 10 contains many false minutiae points while at the threshold value of 20 many genuine minutiae points are rejected. The image depicting the minutiae points at threshold value 15, shows that most of the true minutiae are preserved at this threshold while neglecting the spurious minutiae points. By further analyzing the threshold value between 10-20, it is observed that the quality threshold of 15 gives the best set of true minutiae points. Hence, the quality threshold 15 is selected to determine the eligible minutiae points. After minutiae detection, patches with minutiae points as their center are extracted. We decided the size of the patch to be $96 \times 96$ by performing an experiment over patches of different sizes, that is, $64 \times 64$, $96 \times 96$, and $128 \times 128$. The results of the aforementioned experiment are discussed in section \ref{PSAPatchSize}.


\subsubsection{\textbf{ Section-wise grouping of patches}}
The segmented part of the fingerprint of size $x \times y$ is divided into 9 sections each of size $x/3 \times y/3$ and the minutiae-centered patches are then mapped into these sections based on the location of the minutiae in the segmented image. Patches belonging to the same section are similar in terms of orientations and ridge flow. So mapping the patches section-wise, helps the model to learn salient features precisely. Also, since all the sections are independent, parallel training and testing for section-wise patches would reduce time complexity.

\subsubsection{\textbf{Implementation Details}}
The proposed algorithm is implemented in Python and the models were implemented using the Tensorflow-Keras library. All training and testing were done over NVIDIA TESLA P100 GPU. Each W-GAN was trained for 125 epochs which took around 7-8 hours to converge. The noise input to W-GAN is 128, 1-dimensional random values from the normal distribution. The DenseNet model was trained for 100 epochs which took a maximum of 2 hours to complete its training. To optimize the parameters, the Adam optimizer is used with a learning rate initialized as 0.0001 and weight decay as 0.0004. The optimum threshold values for the global Liveness Score for every model were estimated using the DET curve. Since all the threshold values were very close to 0.5, the threshold value is fixed to be 0.5.

\subsubsection{\textbf{ OPG training and patch generation}}
OPG includes training a W-GAN model associated with every section and for every sensor. For a given dataset with \textit{n} sensors, the OPG consists of $9 \times n$ W-GAN models labeled $WGAN_{ij} (0 \leq i \leq n-1, 0 \leq j \leq 8)$. $WGAN_{ij}$ associated with $i^{th}$ sensor is trained with patches of $j^{th}$ section from all the \textit{n-1} sensors expect the $i^{th}$ sensor to avoid its involvement in the training process.

\begin{figure*}[t!]
	\centering
	\resizebox{0.8\textwidth}{!}{
		{
			
			\includegraphics[]{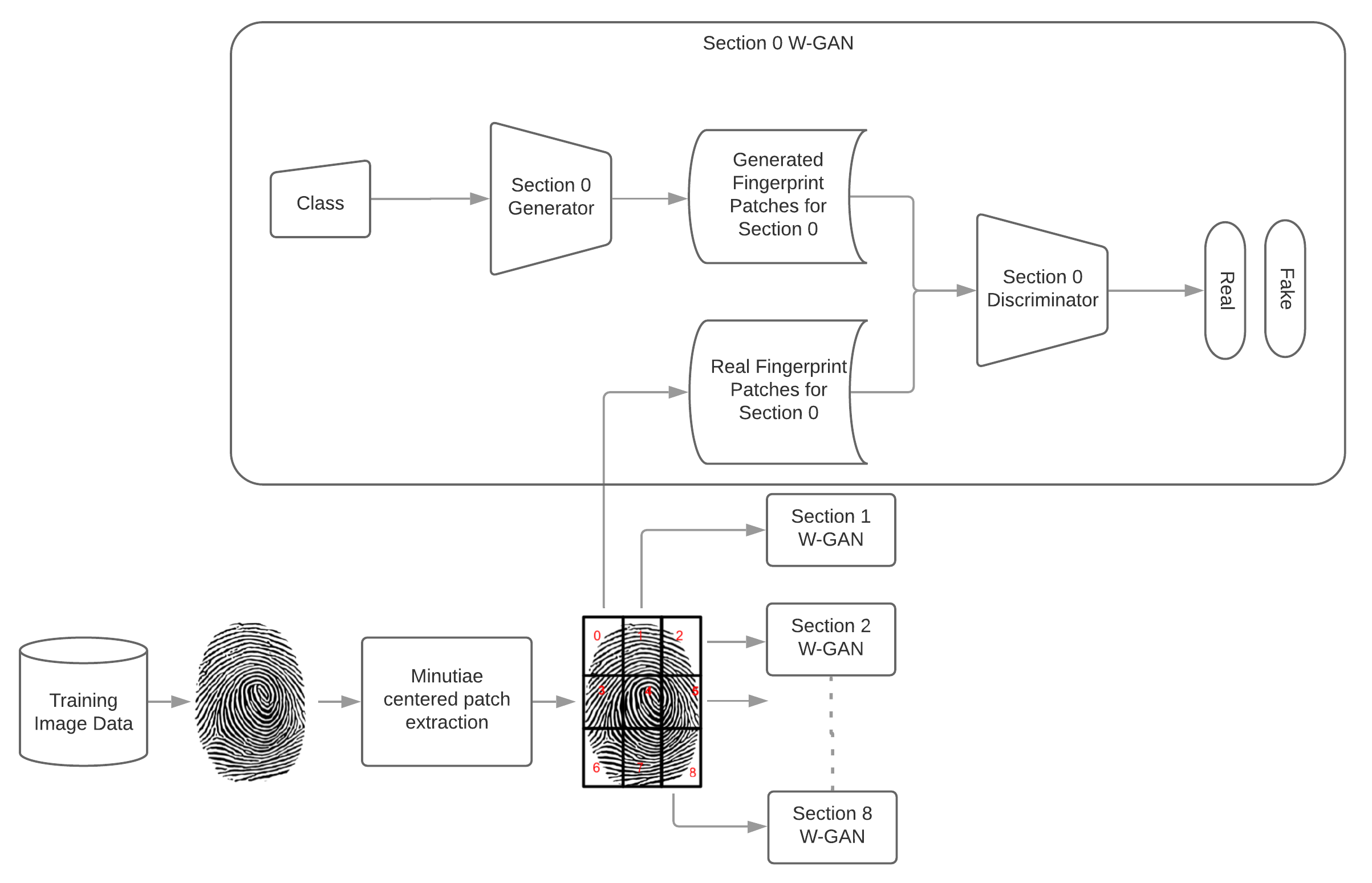}
	}}
	\caption{Block diagram depicting the training process of W-GAN model}
	\label{WGANTrain}
\vspace{3mm}	
\end{figure*}
\begin{figure*}[t]
	\centering
	\includegraphics[width=0.9\textwidth]{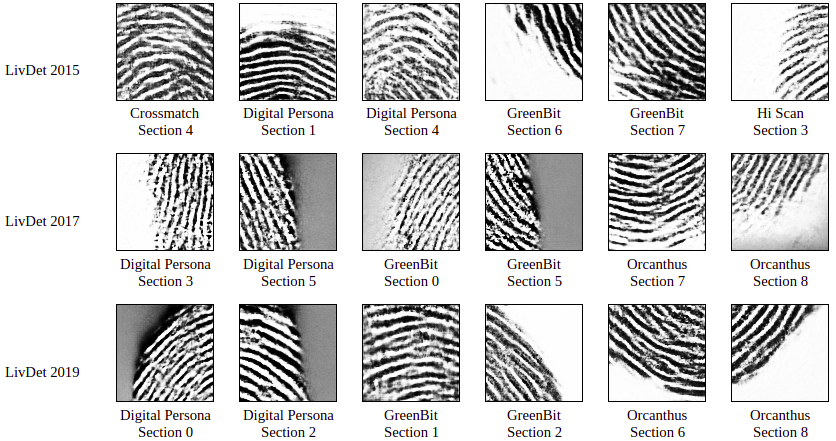}
	\caption{Fingerprint patches generated by the proposed OPG}
	\label{OPGGrid}
\vspace{3mm}	
\end{figure*}
Fig. \ref{WGANTrain} shows the training of the W-GAN with detailed structure of section 0 W-GAN model. W-GAN models for other sections are similar in structure but trained with the patches of respective sections. After training, the generator of the trained models is used to generate synthetic spoof samples. Fig. \ref{OPGGrid} shows some of the patches generated by the proposed OPG for different datasets, sensors, and fingerprint sections.

\subsubsection{\textbf{ Training DenseNet Classifier}}
\begin{figure*}[t]
	\centering
	
	\resizebox{1\textwidth}{!}{
		{
			
			\includegraphics[]{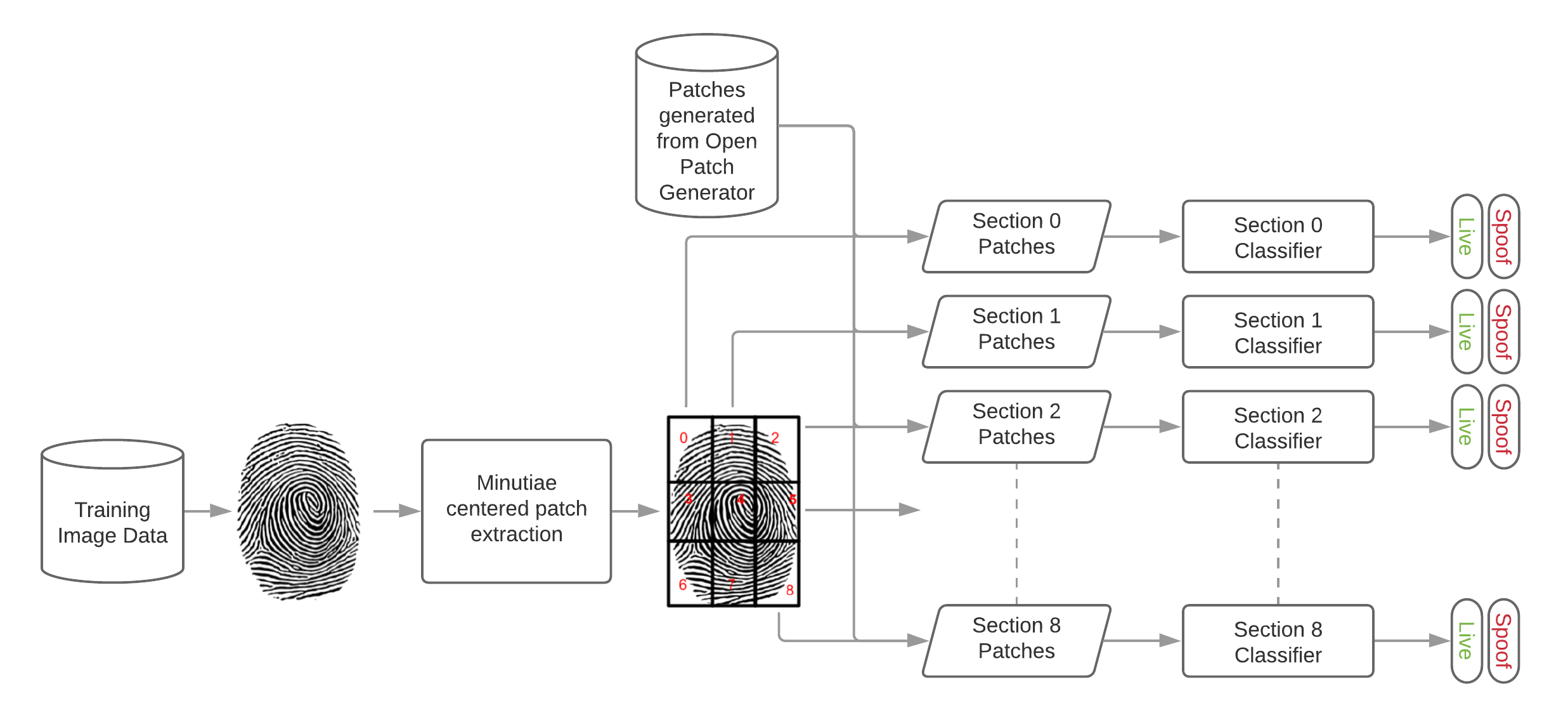}
	}}
	\caption{Block diagram depicting the training process of DenseNet classifier}
	\label{DNetTrain}
 \vspace{-1mm}	
 \end{figure*}

The OPG acts as a wrapper over the spoof detector classifier model. The wrapper feeds the generated synthetic samples to the classifier model by augmenting the samples to training data. This makes the classifier robust against fingerprint samples that might be created using various unknown spoofing materials. CNN DenseNet-121 architecture is used as a classifier. It has been modified for the problem of binary classification because the original classifier is designed for 1000 classes. For the same, a block of 8 layers comprising Pooling, Batch-Normalization, Dropout, Dense, Dense, Batch-Normalization, Dropout, and Dense layers have been added on top of it in the mentioned order. The last Dense layer consist of a single neuron to represent the class of input fingerprints to be either live or a spoof. Finally, the compiled DenseNet models are trained from scratch for each section. The number of synthetic spoof samples provided by the OPG is equal to the number of spoof samples present in the training dataset. The training flow is presented in Fig. \ref{DNetTrain}.
In the figure, the classifier refers to the DenseNet models with the 8 additional layers on top of it.

\section{Experimental Results} \label{experimental res}
\subsection{ \textbf{Database and performance metrics}}
To assess the performance of the proposed approach, experiments are carried out on the Liveness Detection Competition (LivDet) 2015 \cite{b34_livdet2015}, 2017 \cite{b29}, and 2019 \cite{b30} databases.
Each database consists of multiple datasets captured with various sensing devices. The fingerprint samples in all the datasets are arranged separately for the purpose of training and testing. Each dataset comprises fingerprint samples with different quality and resolution as per the configuration of the sensor. The detailed composition of the LivDet databases is provided in Table \ref{tab: Dataset_Details}.\\

The performance of the proposed model is evaluated in accordance with the ISO/IEc IS 30107 standards \cite{misc1}. The values reported for each experiment are the Attack Presentation Classification Error Rate (APCER) denoted with Eq. \ref{APCER} which represents the percentage of misclassified spoof fingerprint images while the Bonafide Presentation Classification Error Rate (BPCER) denoted with Eq. \ref{BPCER} represents the percentage of misclassified live fingerprint images. 
\begin{equation}
\label{APCER}
APCER = \frac{\textnormal{Number of mis-classified fake samples}}{\textnormal{Total fake samples}} \times{100}
\end{equation}

\begin{equation}
\label{BPCER}
BPCER = \frac{\textnormal{Number of mis-classified live samples}}{\textnormal{Total live samples}} \times {100}
\end{equation}
The overall performance of the system is measured by Average Classification Error(ACE) which is an averaged sum of APCER and BPCER. The equation denoting the formulation of ACE is denoted with Eq. \ref{eqn:ACE}.

\begin{equation}
\label{eqn:ACE}
ACE = \frac{APCER + BPCER}{2}
\end{equation}

For live fingerprint patches, the threshold is set to 0.5 according to the binary classification criteria. Thus, a sample having a classification score greater than 0.5 is classified as a live fingerprint while a sample with a classification score less than 0.5 is considered as a spoof one. The ACE can be further utilized to derive the accuracy of the model using the formula mentioned below. \\
\begin{equation}
\label{eqn:ACCURACY}
Accuracy = 100 -  ACE
\end{equation}
 
\begin{table}[t]
\begin{center}
\caption{Database description of LivDet 2015, 2017, and 2019 databases.}
\resizebox{1.0\textwidth}{!}{
\label{tab: Dataset_Details}
\scriptsize 
\begin{tabular}{ p{1.2 cm} p{1.5 cm} p{1.3 cm} p{1.3 cm} p{4.5 cm} }
\hline
\textbf{Database}               & \textbf{Sensor}          & \multicolumn{1}{l}{\textbf{Live}} & \multicolumn{1}{l}{\textbf{Spoof}} & \multicolumn{1}{c}{\textbf{Spoofing Materials}}                                                                        \\ \hline
 \multirow{4}{2 em}{\textbf{\\ \\ \\ \\ LivDet 2015}} & \textbf{CrossMatch}      & 1000/1000                                 & 1473/1448                                 & Body Double, Eco flex, Playdoh, OOMOO, Gelatin                                                                          \\ \cline{2-5} 
                                  &                                 \textbf{Digital Persona} & 1000/1000                                 & 1000/1500                                 & \multirow{3}{3 em}{\begin{tabular}[c]{@{}l@{}}\\ Ecoflex, Latex,  Gelatine,\\ Woodglue, Liquid Ecoflex,\\  RTV\end{tabular}}     \\ \cline{2-4}
                                  &                                 \textbf{GreenBit}        & 1000/1000                                 & 1000/1500                                 &                                                                                                                         \\ \cline{2-4}
                                  &                                 \textbf{HiScan}      & 1000/1000                                 & 1000/1500                                 &                                                                                                                         \\ \cline{1-5} 
                                   \multirow{3}{2 em}{\textbf{\\ \\\ \\ \\ LivDet 2017}} & \textbf{GreenBit}        & 1000/1700                                 & 1200/2040                                 & \begin{tabular}[c]{@{}l@{}}Body double, Ecoflex, Woodglue, \\ Gelatine, Latex, Liquid ecoflex\end{tabular}              \\ \cline{2-5} 
                                  &                                 \textbf{Orcanthus}       & 1000/1700                                 & 1180/2018                                 & \begin{tabular}[c]{@{}l@{}}Body double, Ecoflex, Latex,\\ Woodglue, Gelatine,  Liquid ecoflex\end{tabular}       \\ \cline{2-5} 
                                  &                                 \textbf{Digital Persona} & 999/1692                                  & 1199/2028                                 & \begin{tabular}[c]{@{}l@{}}Body Double, Ecoflex, Woodglue,\\ Gelatine,\\  Latex, Liquid ecoflex\end{tabular}   \\ \cline{1-5} 
                                   \multirow{3}{2 em}{\textbf{\\ \\ \\ \\ LivDet 2019}} & \textbf{GreenBit}        & 1000/1020                                 & 1200/1224                                 & \begin{tabular}[c]{@{}l@{}}Body double, Ecoflex, Woodglue,\\ Mix1, Mix2, Liquid ecoflex \end{tabular}                   \\ \cline{2-5} 
                                  &                                 \textbf{Orcanthus}       & 1000/990                                  & 1200/1088                                 & \begin{tabular}[c]{@{}l@{}}Body double, Ecoflex, Woodglue,\\ Mix1, Mix2, Liquid ecoflex\end{tabular}            \\ \cline{2-5} 
                                  &                                 \textbf{Digital Persona} & 1000/1099                                 & 1000/1224                                 & \begin{tabular}[c]{@{}l@{}}Ecoflex, Gelatine, Woodglue, Latex,\\  Mix1, Mix2, Liquid ecoflex\end{tabular} \\ \hline
\end{tabular}
}

\end{center}

\end{table}
\subsection{\textbf{Patch-Size Selection}}
\label{PSAPatchSize}
\begin{figure*}[!h]
	\centering
	
	\resizebox{0.65\textwidth}{!}{
		{
			
			\includegraphics[]{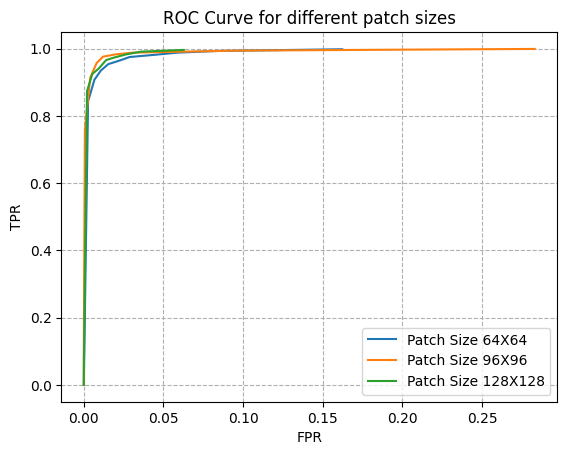}
	}}
	\caption{ROC Curve of the proposed method on fingerprint patches of different size}
	\label{img:PSA_Patch_Size}
 \vspace{-1mm}	
\end{figure*}

The optimal size of the minutiae centered patches are decided using the experiments performed on Greenbit sensor dataset of LivDet 2015. Experimental results using three different patch size of $64 \times 64$, $96 \times 96$, $128 \times 128$ are provided in Table \ref{tab:PSA_Patch_Size}. The proposed model provides the best ACE for the $96 \times 96$ patch size. The performance for the $64 \times 64$ and $128 \times 128$ is significantly lower than the $96 \times 96$ patch size. This is because $64 \times 64$ patch size do not provide an appropriate amount of ridge-valley information and those with size $128 \times 128$ cause overlapping of fingerprint patches. The better performance of $96 \times 96$ patch size can also be seen in the Receiver Operating Characteristic (ROC) curve plotted in Fig. \ref{img:PSA_Patch_Size}. Therefore, we have selected the minutiae centered patch size of $96 \times 96$ for all further experiments.
\begin{table*}[h]
\begin{center}
\caption{Performance of the proposed method on fingerprint patches of different sizes}
\resizebox{1.0\textwidth}{!}{
\label{tab:PSA_Patch_Size}
\begin{tabular}{p{3cm}M{3cm}M{2.5cm}M{2cm}}
\hline
\textbf{Patch Size} & \textbf{BPCER (\%)} & \textbf{APCER (\%)} & \textbf{ACE (\%)} \\ \hline
$64\times64$        & 5.63                & 1.07                & 3.35              \\ \hline
$96\times96$        & 2.21                & 2.34                & 2.27              \\ \hline
$128\times128$      & 5.43                & 0.53                & 2.98              \\ \hline
\end{tabular}
}
\end{center}
\end{table*}

\subsection{\textbf{Testing Procedure}}\label{AA}

The testing of an input fingerprint involves the utilization of trained DensNet models for each of the nine sections. 

\begin{figure*}[!t]
	\centering
	
	\resizebox{0.9\textwidth}{!}{
		{
			
			\includegraphics[]{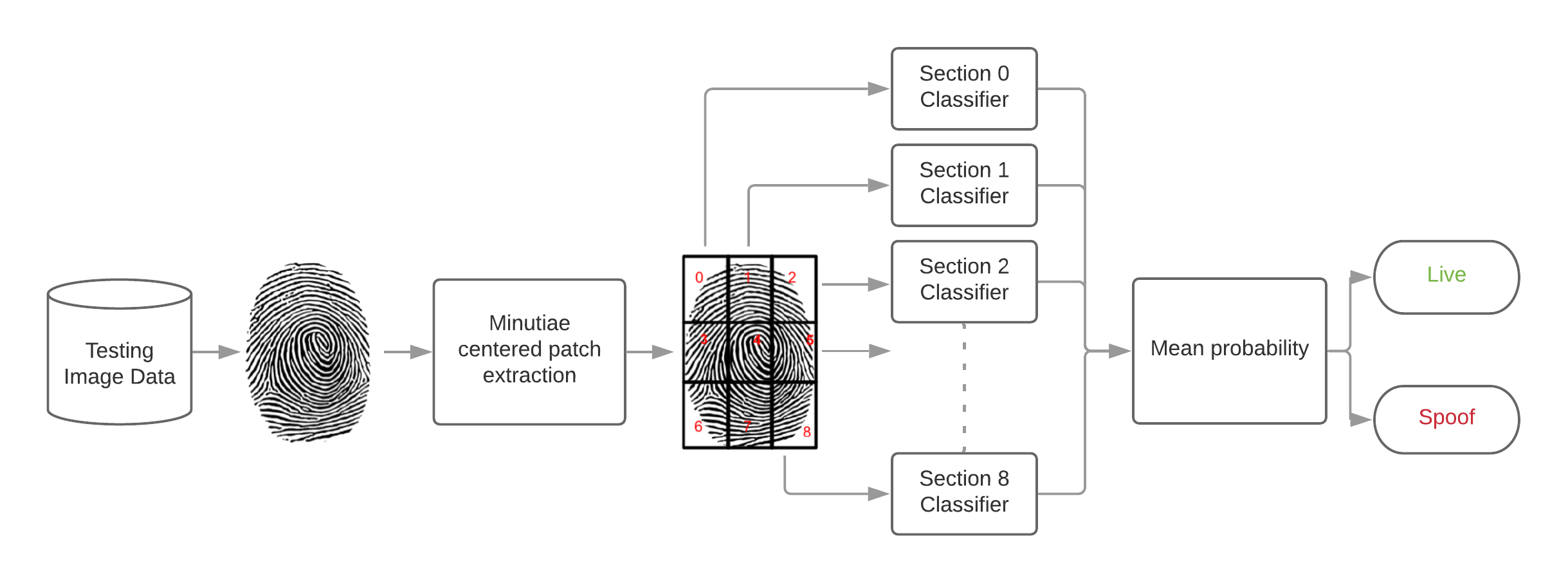}
	}}
	\caption{Block diagram depicting the testing process of DenseNet classifier}
	\label{WGANTest}
 \vspace{-1mm}	
 \end{figure*}

Fig. \ref{WGANTest} shows the testing pipeline. Minutiae-centered patches are extracted from input test images and are mapped to their respective section. The section-wise grouped patches are fed to the trained DenseNet classifier dedicated for a particular section.

The classifier model generates a liveness score for each input patch. The individual liveness score for every patch is then averaged to produce a global liveness score (S) for the entire fingerprint image. The formulation of the global liveness score (S) is denoted with Eq. \ref{liv_score_wgan}.

\begin{equation}
    S = 1/n\sum_{i=1}^{\text{$n$}}s_{i}
    \label{liv_score_wgan}
\end{equation} 
Here, $s_{i}$ is the liveness score of $i^{th}$ patch and n is the total number of patches extracted. An input image is predicted to be live if its global liveness score is more than a threshold value of 0.5 and predicted spoof otherwise.

\vspace{2 mm}
\subsection{\textbf{Comparative Analysis}}\label{compAnalysis}

The performance of the proposed model is assessed in three different scenarios that demonstrate its robustness against samples created with unknown spoof materials and captured with unknown sensors. All the details of these setups are mentioned in the following sections.

\subsubsection{\textbf{Intra-Sensor and Known Spoof Material}}

In this setup, all the training, as well as testing images are acquired with the same sensor and the testing dataset contains the spoof fingerprint samples of the materials which are used during the training. We distributed all the minutiae-centered patches to nine section-wise models which introduce parallel processing and result in less processing time for fingerprints for real-life applications. Table \ref{tab : 2015_Intra_Sensor} presents the performance of the proposed model. Since only LivDet 2015 consists of spoof fingerprint samples created with known and unknown materials, the column ``APCER (Known)'' indicates the performance of the proposed model in this setup. An ACE of 4.61\% is achieved in this setup which indicates its capability to detect the spoofs created with the known materials.

\begin{table*}[h!]
\begin{center}
\caption{Intra-Sensor performance on LivDet 2015 database}
\resizebox{1.0\textwidth}{!}{
\label{tab : 2015_Intra_Sensor}
\vspace{-2mm}
\begin{tabular}{p{1.6cm}M{1.2cm}M{1.2cm}M{1.2cm}M{1.6cm}M{1.0cm}}
\hline
\textbf{Sensor} & \textbf{BPCER (\%)} & \textbf{APCER (\%)} & \textbf{APCER (Known) (\%)} & \textbf{APCER (Unknown) (\%)} & \textbf{ACE (\%)} \\ \hline
CrossMatch      & 2.08          & 2.78         & 3.46                & 1.83                  & 2.49                           \\ \hline
Digital Persona  & 4.42          & 7.87         & 5.48                 & 12.7                   & 6.50                      \\ \hline
HiScan        & 3.71          & 3.89         & 7.5                & 5.62                  & 3.82                           \\ \hline
GreenBit        & 2.21         & 2.34         & 2                 & 3.4                   & 2.27                           \\ \hline
\textbf{Average}    & \textbf{3.10}        & \textbf{4.22}         & \textbf{4.61}                & \textbf{5.88}                  & \textbf{3.80}                          \\ \hline
\end{tabular}
}
\end{center}
\end{table*}

\subsubsection{\textbf{Intra-Sensor and Unknown Spoof Material}}
In this configuration, training and testing samples are collected from the same sensor but the testing dataset have some or all spoof samples belonging to other materials that are not utilized in the training of the model. In this scenario, testing any model stresses the model's performance in a real-world context where a spoof fingerprint may be generated using any novel substance. LivDet 2015 is considered as semi-cross-material dataset since only one-third of testing spoof samples captured by all four sensors are created using unknown material. LivDet 2017 and 2019 are completely cross-material databases. The performance of the proposed model in this evaluation setup is represented by the column labeled ``APCER (Unknown)” for the LivDet 2015 database in Table \ref{tab : 2015_Intra_Sensor}  while the performance over LivDet 2017 and 2019 is reported in Table \ref{tab : 2017_Intra_Sensor} and \ref{tab : 2019_Intra_Sensor}. The ACE of 5.88\%, 5.03\%, and 7.10\% are reported on LivDet 2015, 2017, and 2019 databases, respectively. 
\begin{table*}[]
\begin{center}
\caption{Intra-Sensor Performance on LivDet 2017 database}
\resizebox{1.0\textwidth}{!}{

\label{tab : 2017_Intra_Sensor}
\vspace{-2mm}
\begin{tabular}{p{3cm}M{3cm}M{2.5cm}M{2cm}}
\hline
\textbf{Sensor} & \textbf{BPCER (\%)} & \textbf{APCER (\%)} & \textbf{ACE (\%)} \\ \hline
GreenBit        & 5.02        & 5.47         & 5.26                              \\ \hline
Digital Persona  & 6.09        & 6.58         & 6.36                                \\ \hline
Orcanthus        & 5.37        & 1.87         & 3.47                                 \\ \hline
\textbf{Average}      &       \textbf{5.45}       &      \textbf{4.63}     & \textbf{5.03}                           \\ \hline
\end{tabular}
}
\end{center}
\end{table*}

\begin{table*}[]
\begin{center}
\caption{Intra-Sensor performance on LivDet 2019 database}
\resizebox{1.0\textwidth}{!}{

\label{tab : 2019_Intra_Sensor}
\vspace{-2mm}

\begin{tabular}{p{3cm}M{3cm}M{2.5cm}M{2cm}}
\hline
\textbf{Sensor} & \textbf{BPCER (\%)} & \textbf{APCER (\%)} & \textbf{ACE (\%)} \\ \hline
GreenBit        & 2.65        & 0.16         & 1.29                              \\ \hline
Digital Persona  & 5.89        & 29.0         & 18.45                           \\ \hline
Orcanthus       & 5.05        & 0.09         & 2.45                               \\ \hline
\textbf{Average}    & \textbf{4.52}        &\textbf{10.12}         & \textbf{7.10}                                \\ \hline
\end{tabular}}
\end{center}
\end{table*}

\subsubsection{\textbf{Cross-Sensor}}
In this experimental setup, training and testing images are captured using different sensors. It checks the capability of the model to detect the PAs in case the testing samples are captured with different sensors than the training samples. Table \ref{tab : 2015_Cross_Sensor} lists the outcomes of the model in this scenario for LivDet 2015, while Tables \ref{tab : 2017_Cross_Sensor} and \ref{tab:2019_Cross_Sensor} lists the results for LivDet 2017 and 2019 databases.
\begin{table*}[h!]
\begin{center}
\caption{Cross-Sensor performance on LivDet 2015 database}
\resizebox{1.0\textwidth}{!}{
\vspace{2mm}
\label{tab : 2015_Cross_Sensor}
\begin{tabular}{p{7cm}M{3cm}}
\hline
\textbf{Sensor [Training (Testing)]}            & \textbf{Accuracy (\%)}  \\ \hline
CrossMatch (Digital Persona) & 52.38                                                \\ \hline
CrossMatch (HiScan)         & 57.95                                                 \\ \hline
CrossMatch (GreenBit)        & 59.17                                                \\ \hline
GreenBit (CrossMatch)        & 71.23                                               \\ \hline
GreenBit (Digital Persona)   & 87.47                                               \\ \hline
GreenBit (HiScan)           & 70.44                                              \\ \hline
HiScan (CrossMatch)         & 63.38                                      \\ \hline
HiScan (Digital Persona)    & 81.66                                                \\ \hline
HiScan (GreenBit)           & 85.32                                                \\ \hline
Digital Persona (CrossMatch) & 70.14                                              \\ \hline
Digital Persona (HiScan)    & 60.12                                               \\ \hline
Digital Persona (GreenBit)   & 74.89                                               \\ \hline
\textbf{\textbf{Average}}                      & \textbf{69.51}                         \\ \hline
\end{tabular}
}
\end{center}
\end{table*}

\begin{table*}[h!]
\begin{center}
\caption{Cross-Sensor performance on LivDet 2017 database.} 
\resizebox{1.0\textwidth}{!}{
\begin{tabular}{p{8.0cm}M{3cm}}
\hline
\textbf{\begin{tabular}[c]{@{}c@{}}Sensor \\ Training (Testing)\end{tabular}} & \textbf{\begin{tabular}[c]{@{}c@{}}Accuracy \end{tabular}}  \\ \hline
GreenBit (Orcanthus)                                                          & 58.73                                                                                                           \\ \hline
GreenBit (Digital Persona)                                                     & 88.011                                                                                                          \\ \hline
Orcanthus (GreenBit)                                                           & 50.054                                                                                                          \\ \hline
Orcanthus (Digital Persona)                                                   &  55.699                                                                                                          \\ \hline
Digital Persona (GreenBit)                                                     & 80.31                                                                                                            \\ \hline
Digital Persona (Orcanthus)                                                    & 57.41                                                                                                          \\ \hline
\textbf{Average}                                                              & \textbf{65.09}                                                                                                  \\ \hline
\end{tabular}
}

\label{tab : 2017_Cross_Sensor}
\end{center}
\end{table*}

\begin{table*}[h]
\begin{center}
\caption{Cross-Sensor performance on LivDet 2019 database}
\resizebox{1.0\textwidth}{!}{
\label{tab:2019_Cross_Sensor}
\begin{tabular}{p{7cm}M{3cm}}
\hline
\textbf{Sensor [Training (Testing)]}          & \textbf{Accuracy (\%)} \textbf{\begin{tabular}[c]{@{}c@{}}\end{tabular}} \\ \hline
GreenBit (Orcanthus)        & 58.229                                                                           \\ \hline
GreenBit (Digital Persona)  & 88.141                                                                          \\ \hline
Orcanthus (GreenBit)        & 50.089                                                                           \\ \hline
Orcanthus (Digital Persona) & 56.175                                                                         \\ \hline
Digital Persona (GreenBit)  & 79.679                                                                         \\ \hline
Digital Persona (Orcanthus) & 53.417                                                                        \\ \hline
\textbf{Average}                     & \textbf{64.455}                                                                \\ \hline
\end{tabular}
}
\end{center}
\end{table*}

\subsubsection {\textbf{Comparison with existing approaches in Intra-Sensor paradigm}}
\begin{figure*}[!h]
	\centering
	
	\resizebox{0.75\textwidth}{!}{
		{
			
			\includegraphics[]{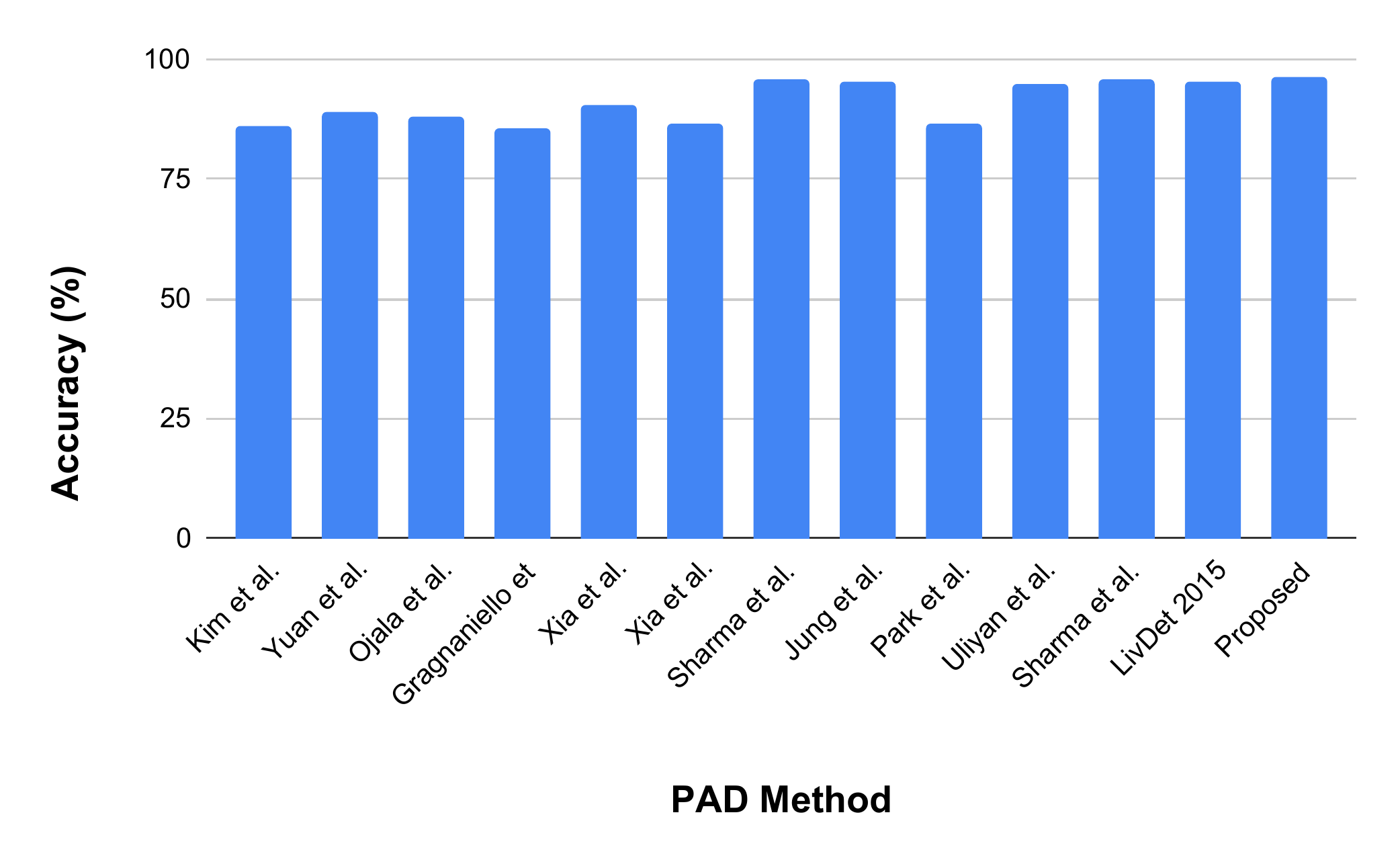}
	}}
	\caption{Comparative analysis of proposed method on LivDet 2015 database in intra-sensor paradigm}
	\label{fig:2015_intra}
 \vspace{-1mm}	
 \end{figure*}

\begin{figure*}[!h]
	\centering
	
	\resizebox{0.75\textwidth}{!}{
		{
			
			\includegraphics[]{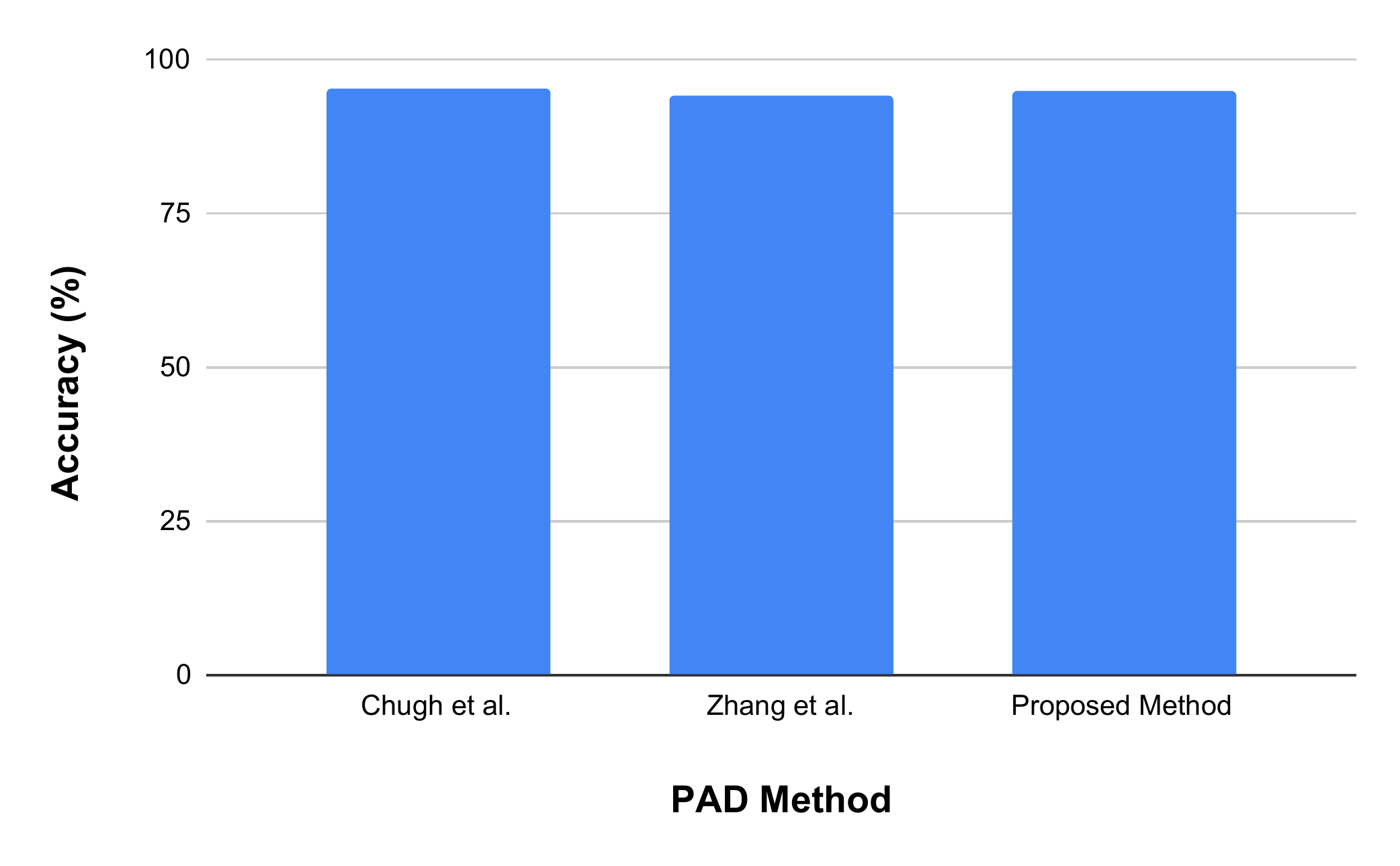}
	}}
	\caption{Comparative analysis of proposed method on LivDet 2017 database in intra-sensor paradigm}
	\label{fig:2017_intra}
 \vspace{-1mm}
 
 \end{figure*}

The obtained results of the proposed approach are compared to state-of-the-art FPAD techniques tested on LivDet 2015, 2017, and 2017 databases. Comparison on LivDet 2015 is presented in Table \ref{tab: Comparison_Intra_2015}. Table \ref{tab: Comparison_Intra_2015} indicates that our proposed method outperforms the methods discussed in \cite{b34}, \cite{b33}, \cite{b35}, \cite{b22}, \cite{b20}, \cite{b13}, \cite{b37}, \cite{b41_park2}, \cite{b4_uliyan}, \cite{b34_livdet2015}, \cite{b5} and \cite{b37}. The aforementioned comparison is also depicted graphically in Fig. \ref{fig:2015_intra}. Similarly, comparison on LivDet 2017 which is mentioned in Table \ref{tab: Comparison_Intra_2017} shows that our proposed method outperforms Chugh et al. \cite{b15} on the Orcanthus dataset, Zhang et al. \cite{b31} on Digital Persona and Orcanthus datasets. The overall performance of our proposed method is comparatively equivalent to the existing works on LivDet 2017. This comparison is also depicted graphically in Fig. \ref{fig:2017_intra}. The result is not compared on LivDet 2019 database since no method is tested on this database in this scenario to the best of our knowledge.

\begin{table*}[t]
\begin{center}

\caption{Comparison with state-of-the-art methods over LivDet 2015 database.}
\resizebox{1.0\textwidth}{!}{
\label{tab: Comparison_Intra_2015}
\begin{tabular}{p{1.8cm}M{1.0cm}M{1.7cm}M{1.5cm}M{1.3cm}M{1.2cm}}
\hline
\textbf{Various Methods}            & \textbf{HiScan} & \textbf{Cross-Match} & \textbf{Green-Bit} & \textbf{Digital Persona} & \textbf {Average} \\ \hline
Kim et al. \cite{b34}& - & -                                                 & -                                                & -                                             & 86.39                        \\ \hline
Yuan et al. \cite{b33}         & -                                                 & -                                                 & -                                                & -                                             & 88.99                        \\ \hline
Ojala et al. \cite{b35}         & -                                                 & -                                                 & -                                                & -                                             & 88.18                        \\ \hline
Gragnaniello et al. \cite{b22} & -                                                 & -                                                 & -                                                & -                                             & 85.73                        \\ \hline
Xia et al. \cite{b20}         & 90.36                                             & 89.18                                             & 95.47                                            & 86.28                                         & 90.33      \\ \hline
Xia et al. \cite{b13}         & -                                             &  -                                           & -                                          & -                                         & 86.39                            \\ \hline
Sharma et al. \cite{b5}         & 95.22                                             &  98.07                                           & 95.7                                          & 94.16                                         & 95.78                            \\ \hline
Jung et al. \cite{b47_jung2}& 95.80                                             &  98.60                                           & 96.20                                          & 90.50                                         & 95.23                            \\ \hline
Park et al. \cite{b41_park2}         &                                              &  -                                           & -                                          & -                                         & 86.39                            \\ \hline
Uliyan et al. \cite{b4_uliyan}         & -                                             &  95.00                                           & -                                          & -                                         & 95.00                            \\ \hline
Sharma et al. \cite{b37}         & 95.5                                             &  95.00                                           & 96.7                                          & 97.2                                         & 96.10                            \\ \hline
LivDet 2015 winner \cite{b34_livdet2015}         & 94.36                                             &  98.10                                           & 95.40                                          & 93.72                                         & 95.39                            \\ \hline

\textbf{Proposed Method}     & \textbf{96.18}                                    & \textbf{97.51}                                    & \textbf{97.63}                                   & \textbf{93.50}                                & \textbf{96.20} \\ \hline
                                                          
\end{tabular}
}
\end{center}

\end{table*}

\begin{table*}[h]
\begin{center}
\label{Comp_2017}    

\caption{Comparison with state-of-the-art methods over LivDet 2017 database.}
\label{tab: Comparison_Intra_2017}
\resizebox{1.0\textwidth}{!}{
\begin{tabular}{p{3.0cm}M{1cm}M{1cm}M{1cm}M{1cm}}
\hline
\textbf{Various Methods} & \multicolumn{1}{l}{\textbf{Orcanthus}} & \multicolumn{1}{l}{\textbf{Digital Persona}} & \multicolumn{1}{l}{\textbf{GreenBit}} & \multicolumn{1}{l}{\textbf{Average}} \\ \hline
Chugh et al. \cite{b15}*              & 94.51                                            & 95.12                                                  & 96.68                                            & 95.43                                 \\ \hline
Zhang et al. \cite{b31}*              & 93.93                                            & 92.89                                                  & 95.20                                            & 94.00                                 \\ \hline
\textbf{Proposed Method}           & \textbf{96.53}                                   & \textbf{93.64}                                         & \textbf{94.74}                                   & \textbf{94.97}                        \\ \hline
\end{tabular}
}
\end{center}
\begin{minipage}{12cm}
\vspace{0.1cm}
\small *The findings of Zhang et al. \cite{b31} and Chugh et al. \cite{b15} are referred from Chugh et al. \cite{b14}.
\hrule

\vspace{0.1cm}

\end{minipage}
\end{table*}

\subsubsection{\textbf{Comparison with existing approaches in Cross-Sensor paradigm}}
\begin{figure*}[!h]
	\centering
	
	\resizebox{0.75\textwidth}{!}{
		{
			
			\includegraphics[]{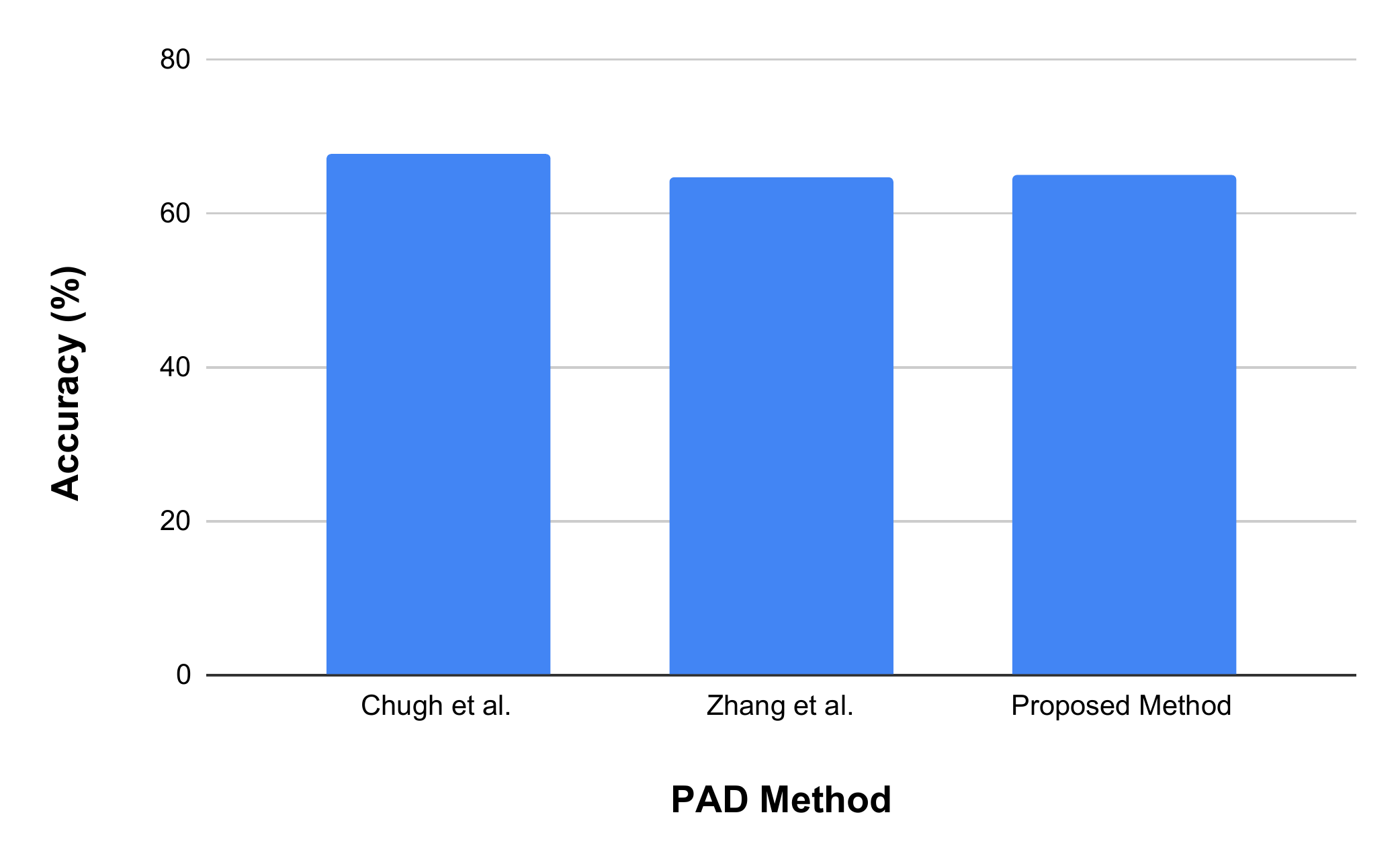}
	}}
	\caption{Comparative analysis of proposed method on LivDet 2017 database in cross-sensor paradigm}
	\label{fig:2017_cross}
 \vspace{-1mm}	
\end{figure*}

A comparative analysis of the cross-sensor experimental performance is given in Table \ref{tab: Comaprison_Cross_2017} which indicates that the proposed model's performance is comparable with \cite{b31} and \cite{b15} on LivDet 2017 database in cross-sensor paradigm. The aforementioned comparison is also depicted graphically in Fig. \ref{fig:2017_cross}. The results of the cross-sensor paradigm on LivDet 2015 and 2019 databases could not be compared as no other existing approach have performed experiments in the cross-sensor experimental setup. The comparative analysis shows that the involvement of the proposed OPG enables the DenseNet classifier to detect PAs in cross-sensor and cross-material paradigms.
\begin{table*}[h!]
\begin{center}
\caption{Comparison with state-of-the-art methods on LivDet 2017 database in cross-sensor paradigm.} 
\resizebox{1.0\textwidth}{!}{
\begin{tabular}{p{4.0cm}M{1.5cm}M{1.5cm}M{1.5cm}}
\hline
\textbf{\begin{tabular}[c]{@{}c@{}}Sensor \\ Training (Testing)\end{tabular}} & \textbf{Zhang et al.} \cite{b31}* & \textbf{Chugh et al.} \cite{b15}* & \textbf{\begin{tabular}[c]{@{}c@{}}Accuracy \end{tabular}}  \\ \hline
GreenBit (Orcanthus)                                                 & 43.98            & 49.43          & 58.73                                                                                                           \\ \hline
GreenBit (Digital Persona)                                           & 80.39            & 89.37          & 88.011                                                                                                          \\ \hline
Orcanthus (GreenBit)                                                 & 68.82            & 69.93          & 50.054                                                                                                          \\ \hline
Orcanthus (Digital Persona)                                          & 62.30            & 57.99          &  55.699                                                                                                          \\ \hline
Digital Persona (GreenBit)                                           & 87.9             & 89.54          & 80.31                                                                                                            \\ \hline
Digital Persona (Orcanthus)                                          & 44.30            & 49.32          & 57.41                                                                                                          \\ \hline
\textbf{Average}                                                             & \textbf{64.61}   & \textbf{67.59} & \textbf{65.09}                                                                                                  \\ \hline
\end{tabular}
}
\label{tab: Comaprison_Cross_2017}
\end{center}
\begin{minipage}{12cm}
\vspace{0.1cm}
\small * The findings of Zhang et al. \cite{b31} and Chugh et al. \cite{b15} are referred from Chugh et al. \cite{b14}.
\hrule

\vspace{0.1cm}

\end{minipage}
\end{table*}

\begin{figure}[h!]
	\centering
	\resizebox{1\textwidth}{!}{
	
	\subfigure
	{	
		\includegraphics[width=0.50\textwidth]{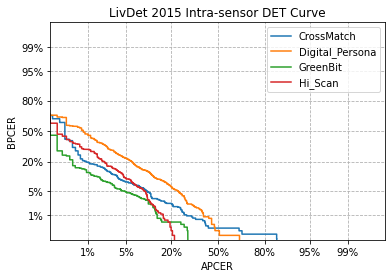}
		\hspace{4 mm}
		\includegraphics[width=0.50\textwidth]{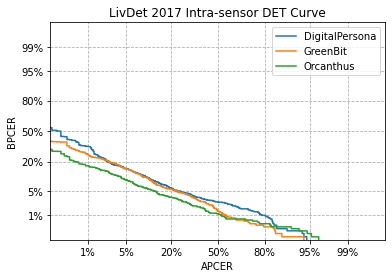}
	}}
	\resizebox{0.60\textwidth}{!}{
	\subfigure
	{
		\includegraphics[width=0.45\textwidth]{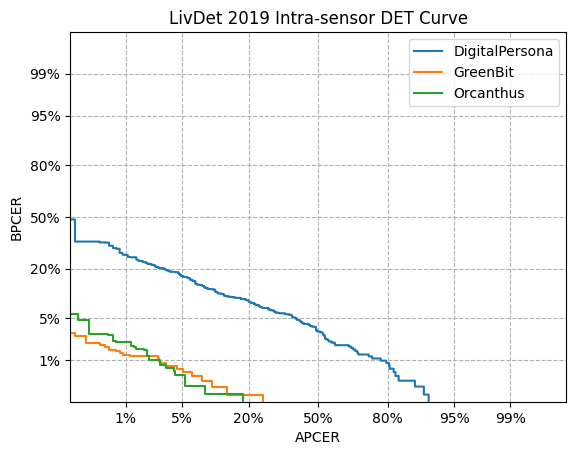}
		\hspace{8 mm}
	}}
	\caption{Detection Error Trade-off (DET) curves for LivDet databases (a) 2015, (b) 2017 and (c) 2019}
	\label{DET_Cur}
	\vspace{-2mm}
	
\end{figure}

We have also reported the Detection Error Trade-off curves for all the datasets of LivDet 2015, 2017, and 2019 databases which are depicted in Fig. \ref{DET_Cur}. It plots APCER vs BPCER and helps in assessing the model's performance. It can be observed that the curves for different sensors of LivDet 2015 and LivDet 2017 are similar and close to each other which shows that the model gives consistent performance across all the sensors in the databases. We can also analyze the high-security performance of the model using the DET curve. For a high-security system, a low APCER is preferred. It can be observed that at a low APCER, around 1\%, for LivDet 2015, the BPCER is around 15 to 30\% for CrossMatch, GreenBit, and HiScan sensors and reaches around 50\% for the Digital Persona sensor. For LivDet 2017, the BPCER is in the range of 20 to 40\% for all sensors in the case of 1\% APCER. For LivDet, 2019 the BPCER is between 1\% and 5\% for Orcanthus and GreenBit, and 50\% for Digital Persona at 1\% APCER.
\vspace{2mm}

\section*{Conclusion}

In this work, we have proposed an OPG wrapper that augments the training dataset with GAN-generated spoof samples which helps in building a generalized PAD model. The training of the DenseNet classifier with the training dataset along with the OPG-generated patches improves its capability to detect spoof fingerprints. It also empowers the classifier to detect fingerprints, created with materials that are unseen at the time of training. The proposed model is validated through a set of experiments using various standard databases which results in improvement over the current state-of-the-art methods. The division of fingerprints into nine sections and processing them in parallel makes it suitable for the various real-time applications required to ensure the security of fingerprint-based recognition systems.\\

\section*{Declaration}
\begin{itemize}
    \item The publicly available databases that are used for the validation of this work can be accessed by the following link:
    https://livdet.org/registration.php
\end{itemize}

\begin{itemize}
    \item Conflict of Interest: The authors declare that they have no conflict of interest
\end{itemize}

\bibliographystyle{sn-basic}
\bibliography{references}


\end{document}